\documentclass{article}

\usepackage{arxiv}

\usepackage[utf8]{inputenc} 
\usepackage[T1]{fontenc}    
\usepackage{hyperref}       
\usepackage{url}            
\usepackage{booktabs}       
\usepackage{amsfonts}       
\usepackage{nicefrac}       
\usepackage{microtype}      
\usepackage{lipsum}		
\usepackage{graphicx}
\graphicspath{{figs/}}
\usepackage{doi}
\usepackage[inline]{enumitem}
\usepackage{xcolor}
\usepackage{fancyvrb}
\usepackage{amsmath}
\usepackage{pifont}
\usepackage{caption}
\usepackage{subcaption}
\usepackage{multirow}
\usepackage{amssymb}

\usepackage{hyperref}
\hypersetup{
    colorlinks=true,
}

\newcommand{\etal}{\textit{et~al.}}

\title{GSV-Cities: Toward Appropriate Supervised \\Visual Place Recognition}

\author{ Amar Ali-bey\\
	Universit\'e Laval\\
	Qu\'ebec, Canada \\
	\And
	Brahim Chaib-draa \\
	Universit\'e Laval\\
	Qu\'ebec, Canada \\
	\And
	Philippe Gigu\`ere \\
	Universit\'e Laval \\
	Qu\'ebec, Canada \\
}

\date{}

\renewcommand{\shorttitle}{GSV-Cities: Toward Appropriate Supervised Visual Place Recognition}

\hypersetup{
pdftitle={GSV-Cities: Toward Appropriate Supervised Visual Place Recognition},
pdfsubject={Visual place Recognition},
pdfauthor={Amar Ali-bey, Brahin Chaib-draa and Philippe Giguere},
pdfkeywords={visual place recognition, place localization, visual geo-localization},
}

\begin{document}
\maketitle

\begin{abstract}
This paper aims to investigate representation learning for large scale  visual place recognition, which consists of determining the location depicted in a query image by referring to a database of reference images. This is a challenging task due to the large-scale environmental changes that can occur over time (i.e., weather, illumination, season, traffic, occlusion). Progress is currently challenged by the lack of large databases with accurate ground truth. To address this challenge, we introduce \textsc{GSV-Cities}, a new image dataset providing the widest geographic coverage to date with highly accurate ground truth, covering more than $40$ cities across all continents over a $14$-year period.
We subsequently explore the full potential of recent advances in deep metric learning to train networks specifically for place recognition, and evaluate how different loss functions influence performance. In addition, we show that performance of existing methods substantially improves when trained on \textsc{GSV-Cities}.
Finally, we introduce a new fully convolutional aggregation layer that outperforms existing techniques, including GeM, NetVLAD and CosPlace, and establish a new state-of-the-art on large-scale benchmarks, such as Pittsburgh, Mapillary-SLS, SPED and Nordland. 
The dataset and code are available for research purposes at \url{https://github.com/amaralibey/gsv-cities}.
\end{abstract}

\vspace{15pt}

\vspace{15pt}
\section{Introduction}
Visual place recognition (VPR) can be defined as the ability for a system to determine whether the location depicted in a query image has already been visited~\cite{arandjelovic2016netvlad}.
This is done by referring to a database of images of previously-visited locations, and comparing the query against them.
VPR has been used in many applications. For instance, it is used in mobile robotics for localization~\cite{engelson1994passive, mur2015orb} and  navigation~\cite{matsumoto1996visual, milford2010persistent}. In particular, it is used to perform loop closure detection in SLAM algorithms~\cite{taketomi2017visual, hane20173d}. 
It is also used in image geo-localization~\cite{weyand2016planet}.

\begin{figure*}[t]
\centering
\includegraphics[width=\linewidth]{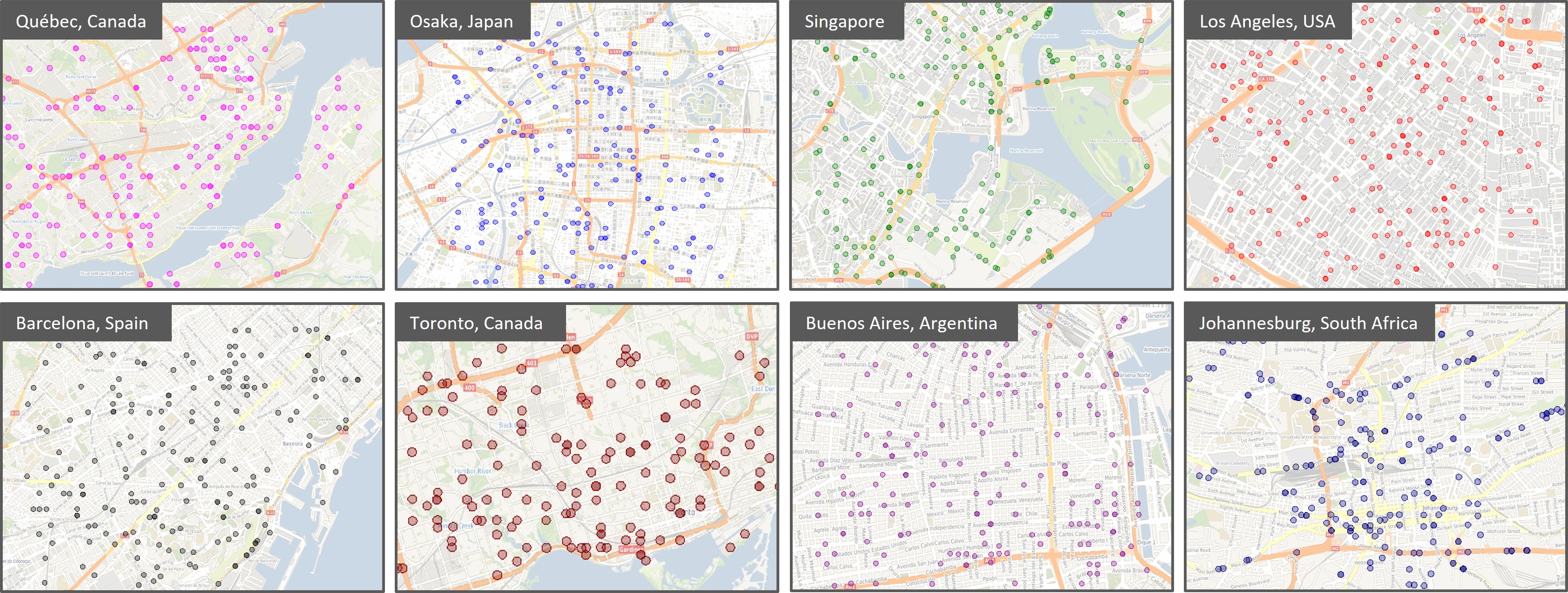}
\caption{Sample locations in $8$ major cities (among 40) in \textsc{GSV-Cities} dataset. All locations are geographically distant and distributed nearly uniformly in every city, maximizing appearance diversity in urban and sub-urban areas. Each location (here, a point) is depicted by at least four images. See details in section~\ref{sec:dataset}.}
\label{fig:geo_sample}
\end{figure*}

Traditional VPR approaches were based on hand-crafted features such as SIFT \cite{lowe2004distinctive}. These can then be further summarised into a single vector representation for the entire image such as Fisher Vectors~\cite{jegou2010aggregating, perronnin2010large}, Bag of Words~\cite{philbin2007object, torii2013visual, galvez2012bags} or VLAD~\cite{jegou2011aggregating,arandjelovic2013all}. With the rise of deep learning, convolutional neural networks (CNNs) \cite{lecun1989backpropagation} showed impressive performance on several computer vision tasks, including image classification \cite{he2016deep}, object detection \cite{liu2020deep}, and semantic segmentation \cite{lateef2019survey}.
For VPR, S\"underhauf~\etal~\cite{sunderhauf2015performance} showed that features extracted from intermediate layers of CNNs trained for image classification can perform better than hand-crafted features. 
As a result, researchers have proposed to train CNNs directly for the task of place recognition~\cite{zhang2021visual}, demonstrating great success at large scale benchmarks~\cite{torii2013visual, warburg2020mapillary}.
However, deep neural networks are data hungry, requiring large amount of training data. This has led to several datasets being released specifically for the training and evaluation of deep neural networks for place recognition. Nevertheless, they all lacked at least one of the following aspects:
\begin{enumerate}[label=(\roman*)]
    \item \textit{Geographical coverage}: most existing datasets are collected in small areas ranging from a city scale \cite{maddern20171} to a small neighborhood \cite{geiger2013vision} or a limited number of locations monitored by a surveillance cameras \cite{chen2017deep} which make them insufficient for training at large scale.
    \item \textit{Accurate ground truth}: except for small-scale datasets, all large-scale ones lack accurate ground truth, which is essential for supervised learning.
    \item \textit{Perceptual diversity}: some datasets do not provide enough appearance variations that can be encountered in real-world applications, such as viewpoint \cite{chen2017deep} or structural changes \cite{olid2018single}.
\end{enumerate}

To the best of our knowledge, there are no datasets that provide precise ground truth, enough environmental diversity and wide geographical coverage, to enable the training of neural networks for place recognition at large scale \emph{and} with full supervision.

The main challenge in training neural networks for place recognition resides in how to learn discriminative representations, such that images depicting the same place get similar representations while those depicting different places get dissimilar ones. So far, most state-of-the-art techniques rely on contrastive or triplet  loss functions to supervise the representation learning process~\cite{zhang2021visual}. This consists of feeding images to the network in form of positive pairs (a pair of images depicting the same place) and negative pairs (a pair of images depicting different places) and optimize the network parameters under a constraint that maximizes the similarity between instances representing the same place or minimizes their similarity in the other case. 

In order to form a negative pair, one can simply choose two distant images (e.g., $50$ meters or more apart) based on their GPS coordinates. However, to form a positive pair, one cannot solely rely on GPS coordinates, as there is no guarantee that the two images are facing the same direction and therefore may not represent the same place.
For instance, Arandjelovic~\etal~\cite{arandjelovic2016netvlad} developed a training procedure, based on a weakly-supervised triplet loss, to enable training from weakly-labeled images. As such, weak supervision is used to compensate for the lack of accurate ground truth in the dataset. Inspired by this work, most recent techniques~\cite{arandjelovic2016netvlad, kim2017learned, liu2019stochastic, ge2020self} relied on weakly-supervised loss functions to train on widely available geotagged images. Despite many advances in visual place recognition, current state-of-the-art techniques are still hindered by the bottleneck of inaccurate ground truth, and thus weak supervision remained their best option.

In this work, we consider representation learning for place recognition as a three components pipeline as shown in Fig.~\ref{fig:pipeline} and introduce the following contributions:
\begin{enumerate}
    \item \textbf{Data module:} we introduce \textsc{GSV-Cities}, a new large-scale dataset with a wide variety of perceptual changes over a 14-year period, covering $40$ cities spread across all continents. This dataset provides \emph{highly accurate ground truth} allowing for straightforward mini-batches sampling eliminating the bottleneck of weak supervision.
    \item \textbf{Image representation:} in addition to \textsc{GSV-Cities} we propose a new fully convolutional aggregation layer (Conv-AP), that generates highly efficient representations while significantly outperforming existing SotA techniques such as GeM~\cite{radenovic2018fine}, NetVLAD~\cite{arandjelovic2016netvlad} and CosPlace~\cite{berton2022rethinking}.
    \item \textbf{Online hard mining and parameters learning:}  given the accurate ground truth of GSV-Cities, we enable online hard samples mining, combined with various SotA metric learning loss functions. By doing so, we show that sophisticated loss functions such as Multi-Similarity~\cite{wang2019multi} can greatly improve performance for visual place recognition.
\end{enumerate}

\section{Related Works}
In this section, we first review existing place recognition methods, then provide a summary of existing datasets that are used for the training and evaluation of such techniques. Finally, we highlight the key issue with weak supervision, which is associated with imprecise ground truth in existing datasets.

\subsection{Place recognition}
The problem of visual place recognition has long been framed as an image retrieval task \cite{arandjelovic2016netvlad, kim2017learned, liu2019stochastic, seymour2019semantically, ge2020self}, where the location of a query image is determined according to the locations of the most relevant images retrieved from a reference database. As for many other computer vision tasks, CNNs trained specifically for place recognition have shown considerable success~\cite{chen2017deep, wang2018omnidirectional, yin2019multi, arandjelovic2016netvlad, ge2020self}. In general, CNNs pre-trained on image classification datasets are adapted to VPR by cropping them at the last convolution layer and plugging a trainable aggregation layer that effectively aggregates feature maps into discriminative representations. 
For instance, Arandjelovic \etal~\cite{arandjelovic2016netvlad} developed an end-to-end trainable version of VLAD descriptor~\cite{arandjelovic2013all}, which can be plugged into a CNN to aggregate deep features into one descriptor. Following the success of NetVLAD, many variants have been introduced. For example, Kim~\etal~\cite{kim2017learned} proposed a method that integrates feature re-weighting into the NetVLAD descriptor. More recently, Yu~\etal~\cite{yu2019spatial} proposed SPE-VLAD, a technique that incorporates pyramid structure into NetVLAD, their aim was to enhance NetVLAD with both spatial and regional features from the images. In another work, Zhang~\etal~\cite{zhang2021vector} proposed two variants, Weighted and Gated NetVLAD, with the latter performing better on VPR tasks. The gating mechanism was applied on each VLAD residual vector to incorporate its personalized characteristics.

Alternative techniques to NetVLAD include R-MAC \cite{tolias2015particular} that consists of extracting Region of Interest (RoI) directly from the CNN feature maps to form representations, and Generalized Mean~(GeM)~\cite{radenovic2018fine} which is a trainable generalized form of global pooling. Recently, Berton~\etal~\cite{berton2022rethinking} introduced CosPlace which combines GeM with a linear projection layer showing great performance boost compared to existing techniques.

In this article, we propose a new efficient aggregation technique that we call Conv-AP. It performs channel-wise pooling on the feature maps followed by spatial-wise adaptive pooling (as described in section~\ref{ssec:conv_ap}), making the architecture fully convolutional and the output dimensionality highly configurable. Conv-AP achieves SotA results on all five benchmarks while being~$16\times$ more compact than NetVLAD.

\subsection{Place recognition datasets}\label{sec:existing_datasets}

\begin{table*}[t]
\centering
\resizebox{\textwidth}{!}{%
\begin{tabular}{|l|rrrrr|ccc|}
\hline
Dataset name                        & Geo. span             & Panoramas           & Places             & Images                   & Time span         & \multicolumn{1}{l}{Accurate GT} & \multicolumn{1}{l}{Viewpoint} & \multicolumn{1}{l|}{Season} \\ \hline \hline
Nordland~\cite{olid2018single}                            & --                    & 0                   & --                  & --                        & 1 year            & {\color{green}\ding{51}}        & {\color{red}\ding{55}}         & {\color{green}\ding{51}}    \\ \hline
SPED \cite{chen2017deep}            & --                    & 0                   & $\sim$2.5K           & 2.5M                     & 7 months            & {\color{green}\ding{51}}        & {\color{red}\ding{55}}         & {\color{green}\ding{51}}    \\ \hline
Oxford RobotCar~\cite{maddern20171}                     & $< 2$ km$^2$         & --                   & --                  & \textbf{20M}             & 18 months         & {\color{green}\ding{51}}        & {\color{green}\ding{51}}       & {\color{green}\ding{51}}    \\ \hline
Pittsburg 250k~\cite{torii2013visual}                      & $\sim 16$ km$^2$     & $\sim$10K           & --                  & 0.25M                    & --                 & {\color{red}\ding{55}}          & {\color{green}\ding{51}}       & {\color{red}\ding{55}}      \\ \hline
TokyoTM~\cite{torii201524}                             & --                    & $\sim$31K           & --                  & 0.19M                     & --                 & {\color{red}\ding{55}}          & {\color{green}\ding{51}}       & {\color{red}\ding{55}}      \\ \hline
Mapillary SLS~\cite{warburg2020mapillary}                       & --                    & --                   & $\sim$23K          & 1.68M                     & 7 years           & {\color{red}\ding{55}}          & {\color{green}\ding{51}}       & {\color{green}\ding{51}}    \\ \hline
\textbf{\textsc{GSV-Cities} (Ours)} & \textbf{2000 km$^2$} & $\sim$\textbf{560K} & \textbf{$\sim$67K} & 0.56M & \textbf{14 years} & {\color{green}\ding{51}}        & {\color{green}\ding{51}}       & {\color{green}\ding{51}}    \\ \hline
\end{tabular}}
\caption{Comparison of datasets for large-scale place recognition.}
\label{tab:datasets}
\end{table*}
Recently, 
several datasets have been released to train and evaluate different techniques of place recognition. In table~\ref{tab:datasets} we summarize some of the relevant ones. \textbf{SPED}~\cite{chen2017deep} was collected from $2.5$k surveillance cameras. It provides accurate ground truth and seasonal changes. However, it is geographically limited and provides no viewpoint changes. \textbf{Nordland}~\cite{olid2018single} was recorded in $4$ seasons with a camera mounted on a train. While this dataset provides accurate ground truth and severe seasonal changes, it lacks viewpoint and urban structural changes. \textbf{Oxford RobotCar}~\cite{maddern20171} contains over $100$ trips of the same $10$ km route in the city of Oxford, UK. It provides highly-accurate ground truth including 3D point cloud. Although this dataset comprises a lot of perceptual changes and contains $20$M images, its geographical coverage is very limited ($<2\,\text{km}^2$) compared to others. \textbf{Pitts250k}~\cite{torii2013visual} and \textbf{TokyoTM}~\cite{torii201524} are among the most used datasets for training and evaluating place recognition techniques. They have been generated from panoramas downloaded from Google Street View. Although they feature significant viewpoint variations and accurate GPS coordinates, they do not provide viewing directions for the images (bearing). Therefore, it is impossible to determine which images depict the same location, based solely on their provided GPS coordinates (positive pairs are almost impossible to form off-the-shelf). Recently, Warburg~\etal~\cite{warburg2020mapillary} introduced \textbf{Mapillary Street Level Sequences (MSLS)}, a large-scale dataset that covers $30$ cities around the world and includes challenging variations in viewpoint, season and illumination. While this dataset provides the viewing direction for each image, it lacks GPS accuracy because the sequences are sourced from Smartphone and Dashcam users. We also note that almost all images in MSLS are forward-facing, causing the road to always appear in the center of the image. Finally, \textsc{\textbf{GSV-Cities}} (ours) includes highly-accurate GPS coordinates \emph{and} viewing direction for each image (see details in section \ref{sec:dataset}) which makes it straightforward to form positive (and negative) pairs.

\subsection{Limitations associated with current datasets: weak supervision}
Current state-of-the art techniques \cite{arandjelovic2016netvlad, kim2017learned, seymour2019semantically, liu2019stochastic, ge2020self, leyva2021generalized, hausler2021patch} rely on datasets of geotagged images for training. However, in such datasets, images are indexed by their GPS coordinates, and split into queries and references. Due to the noisy and weak GPS labels, there is no prior knowledge of the positive matches that are available for each training query. Therefore, most works train the network with weakly supervised loss functions~\cite{arandjelovic2016netvlad} that rely on tuples consisting of a query image, a set of potential positives (images nearby the query according to GPS coordinates) and a set of definite negatives. From an optimization perspective, weakly-supervised loss functions tend to force each query to be closer to its already closest positive (i.e. easiest positive)~\cite{ge2020self}, precluding proper supervision for learning robust feature representations. In this work, we leverage the accurate labels provided by our new dataset to efficiently train networks for place recognition in a supervised manner using existing deep metric loss functions where \emph{difficult positives} are highly effective for learning robust representations~\cite{kaya2019deep}.

\section{Methodology}
We argue that current visual place recognition methods suffer from the bottleneck of weak supervision. To this end, we collect a new dataset that provides highly accurate labels and enables full supervision. Subsequently, we demonstrate, through extensive experiments, that the performance of existing techniques can significantly be increased.

Below we present our data collection approach and an overview of the collected dataset. Then, we propose a new efficient and fully convolutional aggregation layer for visual place recognition.

\subsection{Data collection and overview}\label{sec:dataset}
Following \cite{arandjelovic2016netvlad, torii201524}, we make use of Google Street View Time Machine (GSV-TM) to collect panoramic images depicting the same place over time (between 2007 and 2021). According to \cite{5481932, klingner2013street}, Google Street View combines several types of sensor measurements, including LiDAR data, to correct the position of the collected panoramas. Consequently, the global coordinates provided by Google Street View are of high quality. 
To maximize geographical diversity, we selected $40$ cities spread across all continents. Then, for each city, we uniformly queried GSV-TM in interval of $0.001^{\circ}$ latitude and longitude ($\sim100-130$~meters). This ensured that all locations are physically distant, thus avoiding any overlap (Fig.~\ref{fig:geo_sample} shows location sampled in $8$ different cities). Importantly, only the locations that were visited \emph{at least four times} were retained. 

\begin{figure*}[t!]
\centering
\includegraphics[width=\linewidth]{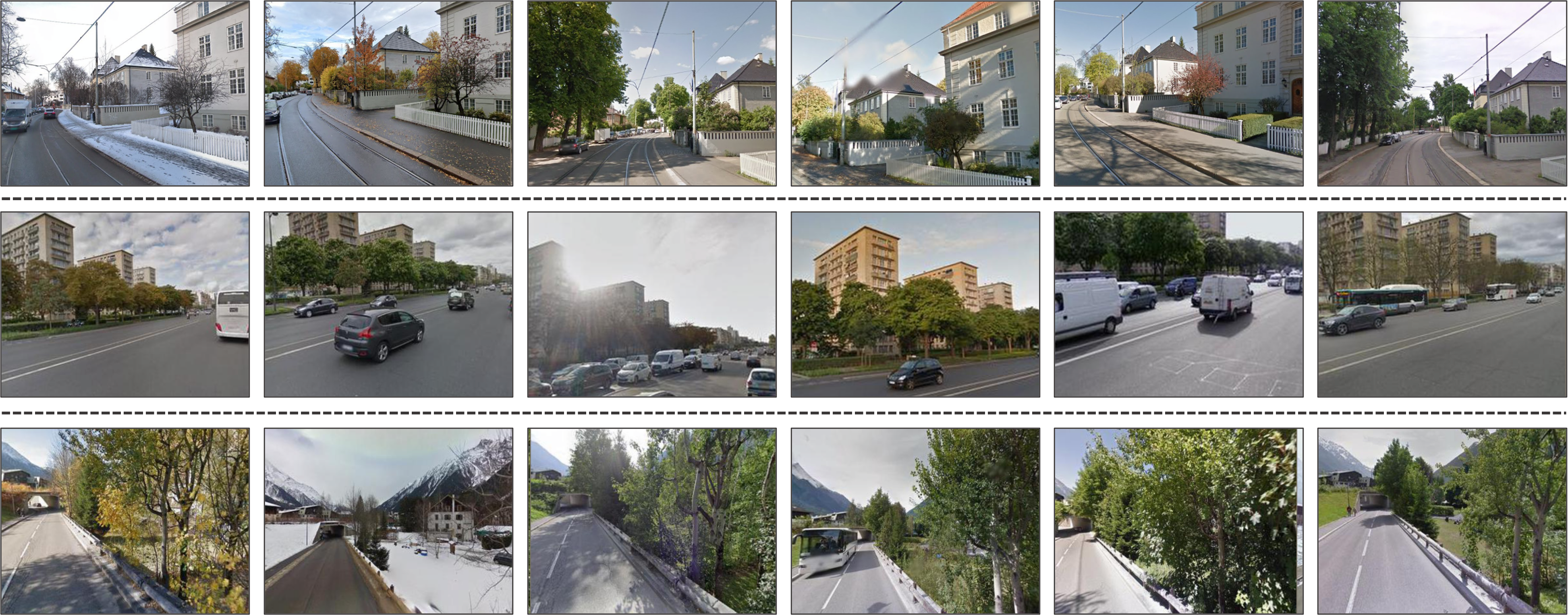}
\caption{Three examples of places in \textsc{GSV-Cities}. Each place is depicted by a set of images (here six, in a row) representing the same physical location. As such, they are indexed by the same ID. The number of images depicting one place in \textsc{GSV-Cities} varies from $4$ up to $20$.}
\label{fig:places_sample}
\end{figure*}

\begin{figure*}[t]%
\begin{subfigure}[Yearly distribution]{0.3\textwidth}
         \centering
         \includegraphics[width=\textwidth]{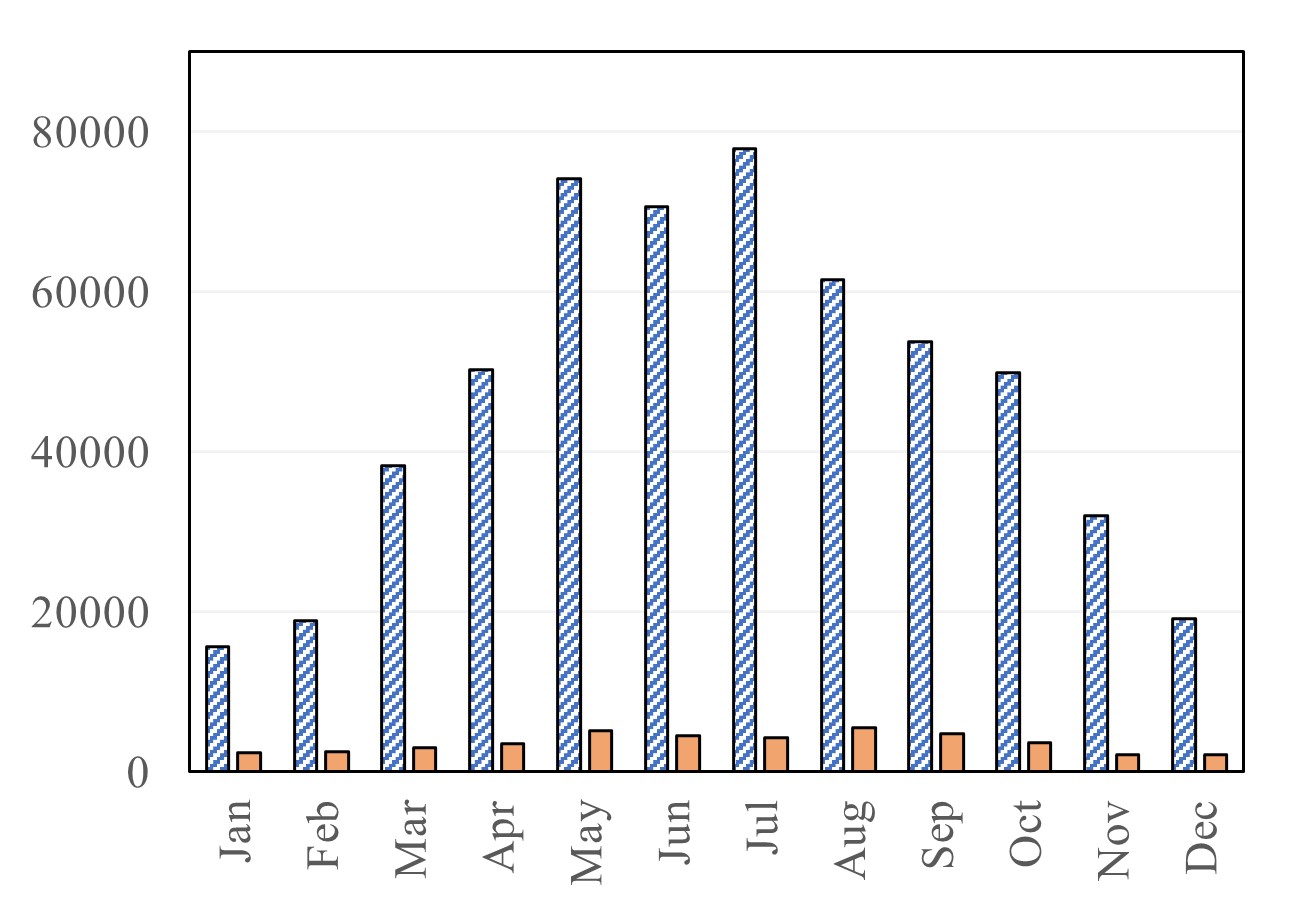}
         \caption{Yearly distribution}
         \label{fig:stats2}
\end{subfigure}
     \hfill
\begin{subfigure}[Monthly distribution]{0.3\textwidth}
         \centering
         \includegraphics[width=\textwidth]{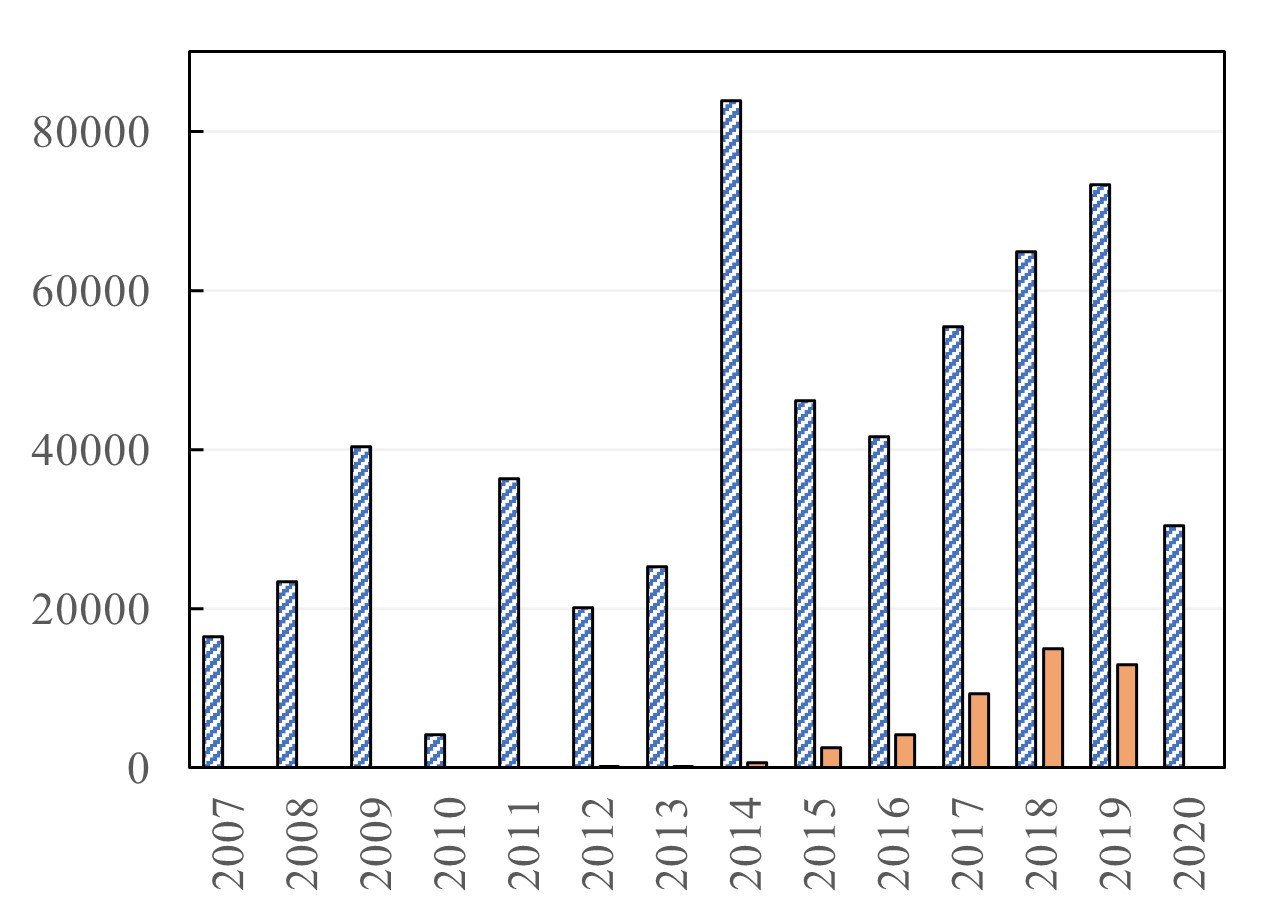}
         \caption{Monthly distribution}
         \label{fig:stats3}
\end{subfigure}
  \hfill
\begin{subfigure}[\# of images per location]{0.3\textwidth}
         \centering
         \includegraphics[width=\textwidth]{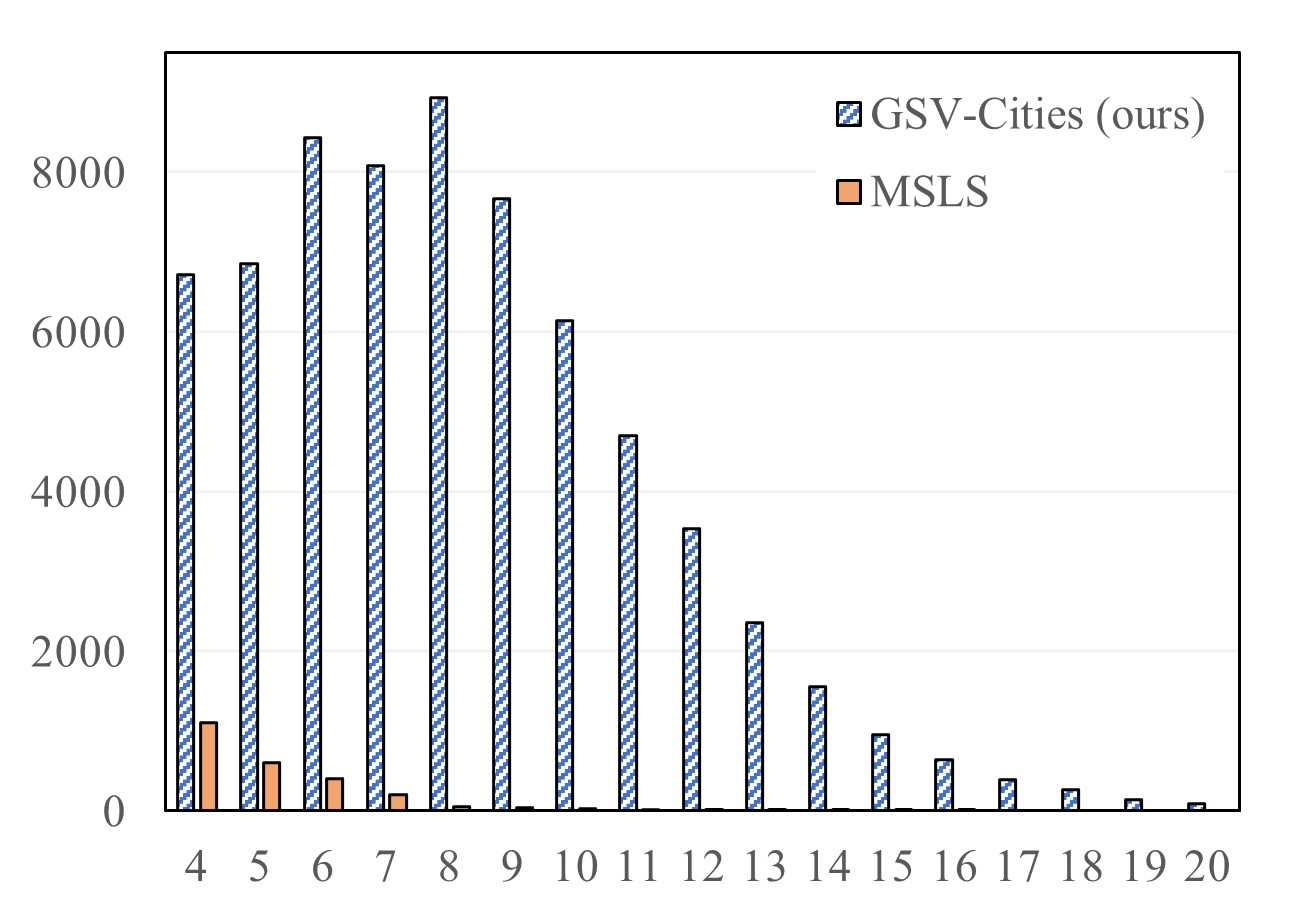}
         \caption{\# of images per location}
         \label{fig:stats1}
\end{subfigure}
\caption{Distribution of images in \textsc{GSV-Cities} (Blue stripped) versus MSLS (Orange), on a yearly and monthly scales. (c) shows the number of images \textit{taken at different dates} in each place (e.g., \textsc{GSV-Cities} contains over $8,000$ places that are depicted by $7$ perspectives each).}
\label{fig:2}
\end{figure*}

We then retrieved each location's historical panoramas and, \emph{very importantly}, their bearings (the direction according to the north pole). This allowed to generate perspectives capturing the exact same place, but at different times (e.g., in Fig.~\ref{fig:places_sample} we show three places, each depicted by $6$ perspectives). We carried out qualitative verification without finding any failure. We compare \textsc{GSV-Cities} (ours) with Mapillary-SLS in~Fig.~\ref{fig:2}.
We show the yearly and monthly distribution of images in Fig.~\ref{fig:stats2} and  Fig.~\ref{fig:stats3} respectively. Clearly, \textsc{GSV-Cities} is far richer and more diverse than MSLS.

\textsc{GSV-Cities} contains over $67,000$k places, each of which is described by a set of images captured at $4$ to $20$ different dates, as can be seen in Fig.~\ref{fig:stats1}. In comparison, MSLS contains $\approx 1,400$ places with images captured at $4$ different dates (versus $6,600$ in \textsc{GSV-Cities}) and $\approx 300$ places with images captured at $6$ different dates (versus $8,300$ in \mbox{\textsc{GSV-Cities}}). This makes \textsc{GSV-Cities} by far the \emph{widest} and \emph{most diverse} dataset in terms of time span and geographical scope. The geographical diversity of \textsc{GSV-Cities} is exemplified in Fig.~\ref{fig:geo_sample}, with locations sampled from $8$ different cities. In total, our dataset covers more than $2,000$~km$^2$.

Finally our dataset is organized as follows: we define a place $P_i$ as a set of images $\left\{I^i\right\}$ depicting the same physical location $P_i = \left( \left\{I_1^i, I_2^i, ..., I_{k_i}^i\right\}, y_i \right)$, where $y_i$ is a an ID assigned to each place. In some sense, $y_i$ can be interpreted as the target label.
Our database $\mathcal{D}$ is then the set of all these geographically-distant places $\left\{P_i\right\}$, namely~$\mathcal{D}~=~\left\{P_1, P_2, ..., P_N\right\}$.


\subsection{Framework of representation learning for visual place recognition}\label{sec:framework}
\begin{figure*}[t]
\centering
\includegraphics[width=\linewidth]{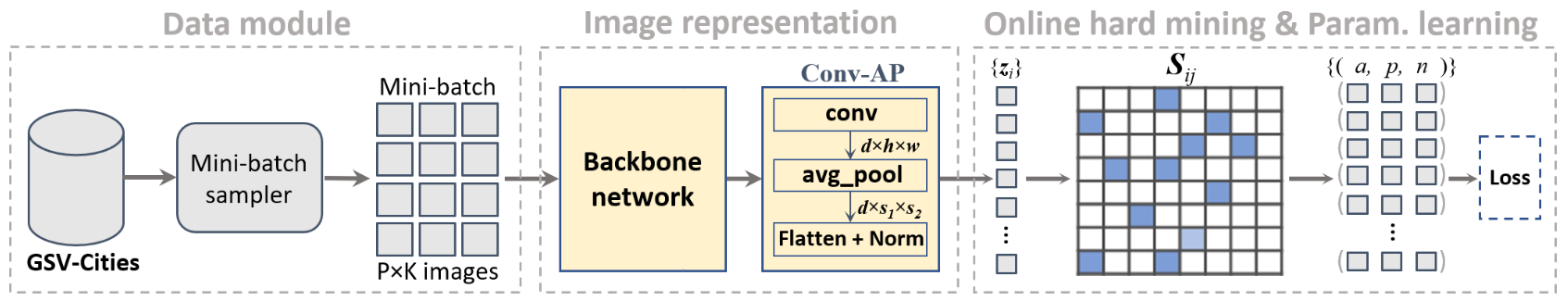}
\caption{Framework of representation learning for visual place recognition. Our dataset, \textsc{GSV-Cities}, makes it straightforward to construct batches comprised of $P$ places each of which depicted by $K$ images. The neural network (backbone + aggregation layer) computes a representation ($\mathbf{z}_i$) for each image in the batch. The matrix $\mathbf{S}$ comprises pairwise similarity between all instances in the batch, which are used to mine informative pairs (or triplets as in this example) in an online fashion. The loss function operates on these pairs/triplets to minimize an objective that maximizes the similarity between instances of the same place and minimizes that of different places.}
\label{fig:pipeline}
\end{figure*}

Inspired by recent advances in deep metric learning~\cite{wang2019multi}, we want to learn place specific representations in a standardized manner.
Our aim is thus to learn a mapping function $\mathit{f_{\mathbf{\theta}}} : \mathcal{D} \subseteq \mathbb{R}^{S} \mapsto \Phi \subseteq \mathbb{R}^{D}$ represented by a neural network (backbone $+$ aggregation layer in Fig.~\ref{fig:pipeline}) parameterized by $\mathbf{\theta}$, which projects images $I_i \in \mathcal{D}$ from a source space $\mathbb{R}^{S}$ (RGB space in our case) into a representation space $\Phi$ where the similarity $S_{ij}$ between two instances $\left(\mathbf{z}_i, \mathbf{z}_j\right)$ is higher if the pair is positive ($\mathbf{z}_i$ and $\textbf{z}_j$ represent the same place) and lower if it is negative ($\mathbf{z}_i$ and $\mathbf{z}_j$ represent two different places). $S_{ij}$ is then defined as:
\begin{equation}
    S_{ij} = \text{sim}\left(\mathit{f_{\theta}}\left(I_i\right), \mathit{f_{\theta}}\left(I_j\right)\right) = \text{sim}\left(\textbf{z}_i, \textbf{z}_j\right)
\end{equation}
where $\text{sim}\left(\cdot\, , \cdot\right)$ represents the cosine similarity. For regularization purposes, the resulting representations are normalized to the real hypersphere $\mathbb{S}^{D}$~\cite{wu2017sampling}, in which the cosine similarity $S_{ij}$ between two instances $\left(\mathbf{z}_i, \mathbf{z}_j\right)$ becomes the inner product of their representations.
With this standardized formulation, we can learn the parameters $\mathbf{\theta}$ (i.e., train the network) using existing pair-based loss functions in a fully supervised fashion by leveraging the accurate labels of \textsc{GSV-Cities}.

\subsection{Fully convolutional feature aggregation}\label{ssec:conv_ap}
Following existing techniques, we use pre-trained networks as backbones cropped at the last convolutional layer.
The output of the backbone is a dense $3$D tensor, called feature maps $\mathbf{F} \in \mathbb{R}^{h \times w \times c}$, where $h {\times} w$ is the spatial resolution and $c$ is the number of channels, which corresponds to the numbers of filters in the last convolutional layer of the backbone. $\mathbf{F}$ can be interpreted as a set of $c$-dimentional descriptors $\mathbf{f}_{ij} \in \mathbf{F}$ at each $h {\times} w$ spatial location.

We introduce a new fully convolutional feature aggregation technique, called Conv-AP that operates as follows. First, given a feature maps $\mathbf{F}$, we apply a channel-wise weighted pooling that linearly projects each spatial descriptor $\mathbf{f}_{ij} \in \mathbf{F}$ into a compact $d$-dimensional representation space, enabling dimensionality reduction along the channel dimension. This can be implemented using a $1 {\times} 1$ convolution with parameters $\mathbf{W} \in \mathbb{R}^{d \times c \times 1 \times 1}$ such as:
\begin{equation}
    \mathbf{F'} = \mathbf{W} \circledast \mathbf{F} = \verb!Conv!_{1 \times 1} (\mathbf{F})
\end{equation}
where $\circledast$ is the convolution operation and $\mathbf{F'} \in \mathbb{R}^{h \times w \times d}$ is the resulting feature maps with depth $d$.
Second, we reduce the spatial dimensionality of $\mathbf{F'}$ using adaptive average pooling (\verb!AAP!), which spatially splits the feature maps into $s_1 {\times} s_2$ equal sub-regions and takes their average, resulting in feature maps of size $s_1~{\times}~s_2~{\times}~d$. In other words, \verb!AAP! effectively determines the stride and window size of the average pooling operation, to obtain feature maps of fixed spatial size. Note that global average pooling~\cite{lin2013network} is a special case of \verb!AAP! where $s_1 = s_2 = 1$. 

Formally, our technique, Conv-AP, can be summarized as follows. Given an image $I_i$, we pass it through the pre-trained backbone to obtain its feature maps $\mathbf{F}_i$, and pass them to the aggregation layer to obtain the final representation $\mathbf{z}_i$, such as:
\begin{equation}
    \mathbf{z}_i =  \verb!AAP!_{s_1 \times s_2} (\, \verb!Conv!_{1 \times 1} (\,\mathbf{F}_i\, ) \,)
\end{equation}
Finally, the resulting representation $\mathbf{z}_i \in \mathbb{R}^{s_1 \times s_2 \times d}$ is flattened and $L_2$-normalized, in order to conform to the training framework in section~\ref{sec:framework}.

\subsubsection{Loss function}\label{sec:loss}

Existing VPR techniques~\cite{arandjelovic2016netvlad, kim2017learned, liu2019stochastic, warburg2020mapillary} train the network in a weakly supervised manner by feeding it tuples $\left(q, \mathbf{P}^q, \mathbf{N}^q \right)$, each of which consists of a query $q$, a subset of potential positives $\mathbf{P}^q$ and a subset of hard negatives $\mathbf{N}^q$. The images in $\mathbf{P}^q$ are geographically close to the query $q$ ($\leq 10$ meters), but they don't necessarily depict the same place (as they might not have the same orientation as $q$). In this case, weakly supervised triplet loss~\cite{arandjelovic2016netvlad} is used, where only the easiest positive from $\mathbf{P}^q$ is used to form a positive pair, as such: 
\begin{equation}
    \mathcal{L}_{\text{w-triplet}} = \sum_{n_j \in \mathbf{N}^q} \left[ \overbrace{\text{sim} \left(q, n_{j}\right) }^{\text{negative pairs}}
     \; - \; \overbrace{\underset{p_i \in \mathbf{P}^q}{\max}  \; \; \text{sim} \left(q, p_{i} \right)}^{\text{the positive pair}} \; + \; m  \right]_+
\label{loss:w_triplet}
\end{equation}
where $m$ is a margin used to exclude trivial triplets and $[\bullet]_+~=~\max(\bullet, 0)$ is the hinge function. Intuitively, from all images in $\mathbf{P}^q$ the one most similar to $q$ is the most likely to depict the same place as the query, and is hence used as the positive pair.

In this work, we follow a different approach by taking advantage of the accurate labels provided by \textsc{GSV-Cities} and adopt a formulation similar to~\cite{hermans2017defense}, where batches are formed by sampling $P$ places ($P$ different IDs) from the dataset, and randomly picking $K$ images of each place (\textsc{GSV-Cities} guarantees that the $K$ images will depict the same place), thus resulting in batches of size $P {\times} K$ (as seen in Fig.~\ref{fig:pipeline}). This formulation allows us to use pair-based and triplet-based loss functions in a supervised manner such as (1)~Contrastive loss~\cite{hadsell2006dimensionality}, (2)~Triplet ranking loss~\cite{hoffer2015deep} and (3)~Multi-Similarity loss~\cite{wang2019multi}.

\vspace{2pt}
\noindent\textbf{Contrastive loss:} this function~\cite{hadsell2006dimensionality} aims at maximizing the similarity between positive pairs and minimizing it for negative pairs as follows:
\begin{equation}
    \mathcal{L}_{\text{contrast}} = \left(1-\mathcal{I}_{i j} \right) \left[S_{ij} - m \right]_{+} - \mathcal{I}_{i j} S_{i j}
\label{loss:contrastive}
\end{equation}
where $\mathcal{I}_{i j} = 1$ indicates a positive pair, and $0$ otherwise. The margin $m$ avoids optimizing indefinitely the negative pairs that are already dissimilar enough.

\vspace{2pt}
\noindent\textbf{Triplet margin loss:} this function~\cite{hoffer2015deep} can be calculated on a triplet $\left(\mathbf{z}_i, \mathbf{z}_j, \mathbf{z}_k\right)$ as follows:
\begin{equation}
\mathcal{L}_{\text{triplet}} = \left[S_{ij} - S_{ik} + m \right]_{+}
\label{loss:triplet}
\end{equation}
where the objective is to reduce the similarity $S_{ik}$ of the negative pair $\left(\mathbf{z}_{i}, \mathbf{z}_{k}\right)$ and at the same time increase the similarity $S_{ij}$ of the positive pair  $\left(\mathbf{z}_{i}, \mathbf{z}_{j}\right)$.

\vspace{2pt}
\noindent \textbf{Multi-Similarity loss:} Wang~\etal~\cite{wang2019multi} recently introduced this loss function which incorporates advanced pair weighting schemes and can be calculated as follows:
\begin{equation}
\small
\begin{aligned}
\mathcal{L}_{\text{MS}} = \frac{1}{N} \sum_{i=1}^{N}  \left\{  \frac{1}{\alpha} \log \left[ 1 + \sum_{j \in \mathcal{P}_{i}} e^{- \alpha \left(S_{ij} - m \right)} \right]  +    \frac{1}{\beta}  \log \left[ 1+\sum_{k \in \mathcal{N}_{i}} e^{\beta \left(S_{ik} - m \right)} \right] \right\}
\end{aligned}
\label{loss:msloss}
\end{equation}
where for each instance $\mathbf{z}_i$ in the batch, $\mathcal{P}_{i}$ represents the set of indices $\{j\}$ that form a positive pair with $\mathbf{z}_i$ and $\mathcal{N}_{i}$ the set of indices $\{k\}$  that form a negative pair with $\mathbf{z}_i$. $\alpha, \beta \text{ and } m$ represent constants (hyperparameters) that control the weighting scheme (refer to~\cite{wang2019multi} for more details).

Finally, there are many other pair/triplet based loss functions in metric learning literature~\cite{musgrave2020pytorch, kaya2019deep} that we can use for training, such as FastAP~\cite{cakir2019deep} and Circle loss~\cite{sun2020circle}.

\subsubsection{Online hard mining}\label{sec:sampling}
Current techniques that rely on weakly supervised triplet loss~\cite{arandjelovic2016netvlad, kim2017learned, liu2019stochastic, warburg2020mapillary} employ offline hard negative mining, which consists of retrieving, for each query, the most difficult negatives among all (or a subset of) images in the dataset. This is done to compensate for the drawback of easy positives used in the weakly supervised loss function (Eq.~\ref{loss:w_triplet}). Nonetheless, offline mining is computationally expensive where the training is paused and representations of a large subset of images are computed and stored into a cache memory.

In this work, we are using batches of size $P {\times} K$, where each sample can play the role of query, positive and negative at the same time. This formulation can generate a large number of pairs (or triplets), many of which can be uninformative. Yet choosing informative positive and negative pairs within the batch is crucial to robust feature learning. An informative pair is one that produces a large loss value, thereby pushing the network to learn discriminative representations. Thus, an effective mining strategy is able to increase not only performance, but also the training speed~\cite{kaya2019deep}. We opt for online mining which, unlike offline mining, does not induce a lot of computation since it is performed on the fly on each batch at the end of the forward pass (as shown in Fig.~\ref{fig:pipeline}). Many online mining strategies have been proposed in the literature~\cite{musgrave2020pytorch}. For example, Online Hardest Mining (OHM) strategy \cite{hermans2017defense} consists of keeping only the most difficult positive and the most difficult negative for each instance in the batch. Recently, Wang~\etal~\cite{wang2019multi} proposed a pair mining strategy that considers multiple pairwise similarities in the mining process, demonstrating great boost of performance.

In summary, the proposed framework (as shown in Fig.~\ref{fig:pipeline}) makes it straightforward to train neural networks for place recognition without resorting to weak supervision and offline hard example mining, while also accelerating training time by orders of magnitude and improving performance of existing techniques as we show in section~\ref{sec:exp:netvlad}.

\section{Experiments}
In this section we describe the datasets and the metrics used for training and evaluation (section \ref{sec:exp:eval}), and then show how training with \textsc{GSV-Cities} improves performance of existing techniques (section \ref{sec:exp:netvlad}). Next, we evaluate how different metric loss functions~\cite{musgrave2020pytorch} perform for training place recognition networks (section \ref{sec:exp:metric_learning}). We then compare our new aggregation method (Conv-AP) to existing state-of-the-art techniques on multiple large-scale benchmarks (\mbox{section \ref{sec:exp:sota}}). We also show some extended implementation details (\ref{sec:exp:extended}) and experiment how different backbone architectures affect performance (section \ref{sec:exp:backbone}). Finally, we show the impact of applying PCA dimensionality reduction (section \ref{sec:exp:pca}).

\subsection{Datasets and metrics}\label{sec:exp:eval}
For evaluation, we use the following $4$ benchmarks,  Pitts250k-test~\cite{torii2013visual}, MSLS~\cite{warburg2020mapillary}, SPED~\cite{zaffar2021vpr} and Nordland~\cite{zaffar2021vpr}. They contain respectively, $8$k, $750$, $607$ and $1622$ query images, and $83$k, $19$k, $607$ and $1622$ reference images. We follow the same evaluation metric of \cite{arandjelovic2016netvlad, warburg2020mapillary, zaffar2021vpr}, where the Recall@k is measured. For Pitts250k and MSLS benchmarks, the query image is determined to be successfully retrieved if at least one of the top-$k$ retrieved reference images is located within $d = 25$ meters from the query (according to their GPS coordinates).

\subsubsection{Implementation details}
By default, we use ResNet50~\cite{he2016deep} as backbone pre-trained on ImageNet~\cite{russakovsky2015imagenet} and cropped at the last residual bloc, extended with an aggregation layer. For NetVLAD~\cite{arandjelovic2016netvlad} we fix the number of clusters to $16$, resulting in $32$k-dimensional representations as in \cite{arandjelovic2016netvlad, hausler2021patch}. Unless otherwise stated, for our method (Conv-AP) we fix the depth of the channel-wise pooling operation (the $1 {\times} 1$ convolution) to $d = 2048$.

For data organization, we use batches of size $P = 100$ places, each one being represented by $K = 4$ images, resulting in batches of size $400$. Stochastic gradient descent (SGD) is utilized for optimization, with momentum $0.9$ and weight decay $0.001$. The initial learning rate of $0.03$ is multiplied by $0.3$ after every $5$ epochs.  We train for a maximum of $30$ epochs using images resized to $320{\times} 320$.

\subsection{Importance of training with GSV-Cities}
\label{sec:exp:netvlad}
To assess the benefits of training with our dataset, we compare the performance of three existing methods when trained on either Pitts30k-train~\cite{torii2013visual}, MSLS-train~\cite{warburg2020mapillary} or \textsc{GSV-Cities}.

\setlength{\tabcolsep}{10pt}
\begin{table}
\caption{Training on \textsc{GSV-Cities} versus MSLS and Pittsburg. Performance of AVG (global average pooling), GeM (Generalized Mean) and NetVLAD are reported.}
\resizebox{\textwidth}{!}{%
\begin{tabular}{@{}ll c@{\hspace{5pt}}c@{\hspace{5pt}}c c@{\hspace{5pt}}c@{\hspace{5pt}}c c@{\hspace{5pt}}c@{\hspace{5pt}}c c@{\hspace{5pt}}c@{\hspace{5pt}}c@{}}
\toprule
\multirow{2}{*}{Method}  & \multirow{2}{*}{Training dataset} & \multicolumn{3}{c}{Pitts250k-test}            & \multicolumn{3}{c}{MSLS-val}                  & \multicolumn{3}{c}{SPED}                      & \multicolumn{3}{c}{Nordland}                  \\ \cmidrule(l{0.3cm} r{0.3cm}){3-5} \cmidrule(l{0.3cm} r{0.3cm}){6-8} \cmidrule(l{0.3cm} r{0.3cm}){9-11} \cmidrule(l{0.3cm} r{0.3cm}){12-14} 
                         &                                   & \footnotesize{R@1}           & \footnotesize{R@5}           & \footnotesize{R@10}          & \footnotesize{R@1}           & \footnotesize{R@5}           & \footnotesize{R@10}          & \footnotesize{R@1}           & \footnotesize{R@5}           & \footnotesize{R@10}          & \footnotesize{R@1}           & \footnotesize{R@5}           & \footnotesize{R@10}          \\ \midrule
\multirow{2}{*}{AVG}     & MSLS~\cite{warburg2020mapillary}  & 62.6          & 82.7          & 88.4          & 59.3          & 71.9          & 75.5          & 54.7          & 72.5          & 77.1          & 4.4           & 8.4           & 10.4          \\
                         & GSV-Cities (ours)                 & \textbf{78.3} & \textbf{89.8} & \textbf{92.6} & \textbf{73.5} & \textbf{83.9} & \textbf{85.8} & \textbf{58.8} & \textbf{77.3} & \textbf{82.7} & \textbf{15.3} & \textbf{27.4} & \textbf{33.9} \\ \midrule
\multirow{2}{*}{GeM~\cite{radenovic2018fine}}     & MSLS~\cite{warburg2020mapillary}                              & 72.3          & 87.2          & 91.4          & 65.1          & 76.8          & 81.4          & 55.0          & 70.2          & 76.1          & 7.4           & 13.5          & 16.6          \\
                         & GSV-Cities (ours)                 & \textbf{82.9} & \textbf{92.1} & \textbf{94.3} & \textbf{76.5} & \textbf{85.7} & \textbf{88.2} & \textbf{64.6} & \textbf{79.4} & \textbf{83.5} & \textbf{20.8} & \textbf{33.3} & \textbf{40.0} \\ \midrule
\multirow{3}{*}{NetVLAD~\cite{arandjelovic2016netvlad}} & Pitts30k                          & 86.0          & 93.2          & 95.1          & 59.5          & 70.4          & 74.7          & 71.0          & 87.1          & 90.4          & 4.1           & 6.6           & 8.2           \\
                         & MSLS~\cite{warburg2020mapillary}  & 48.7          & 70.6          & 78.9          & 48.6          & 63.4          & 70.5          & 37.9          & 56.0          & 64.9          & 2.4           & 5.0           & 6.6           \\
                         & GSV-Cities (ours)                 & \textbf{90.5} & \textbf{96.2} & \textbf{97.4} & \textbf{82.6} & \textbf{89.6} & \textbf{92.0} & \textbf{78.7} & \textbf{88.3} & \textbf{91.4} & \textbf{32.6} & \textbf{47.1} & \textbf{53.3} \\ \bottomrule
\end{tabular}%
}
\label{netvlad}
\end{table}

Table~\ref{netvlad} reports the performance of NetVLAD~\cite{arandjelovic2016netvlad}, GeM~\cite{radenovic2018fine} and AVG (which represents global average pooling), trained on Pitts30k, MSLS or \textsc{GSV-Cities}. As we can see, using \textsc{GSV-Cities} for training drastically improves performance of all three techniques across all benchmarks.
Training AVG on \textsc{GSV-Cities} instead of MSLS improved its recall@1 performance (in percentage points) by $15.7$ on Pitts250k, $14.2$ on MSLS, $4.1$ on SPED and $10.9$ on Nordland. Note that the relative improvement on Nordland is $240\%$.
We also report the performance of training GeM on \textsc{GSV-Cities} versus MSLS, where its recall@1 improved by, respectively, $10.6,  \, 10.4, \, 9.4$ and $13.4$ percentage points, which is clearly a significant boost of performance.

Most interestingly, NetVLAD drastically increases in performance when trained on \textsc{GSV-Cities}, ($86.0\% \rightarrow 90.5\%$) on Pitts250k , ($59.5\% \rightarrow 82.6\%$) on MSLS-val, and ($4.1\% \rightarrow 32.6\%$) on Nordland. This highlights the importance of the accurate ground truth of our dataset.
Note that when NetVLAD is trained on MSLS, it reaches convergence after \mbox{\textbf{55 days}} of training (reported by the authors of~\cite{hausler2021patch}), compared to \textbf{8 hours} of training on \textsc{GSV-Cities}, which translates to $99.4\%$ less training time (in other words, \textit{it takes less time to train NetVLAD 165 times on \textit{GSV-Cities} than to train it once on MSLS}). This large difference is due to the fact that using MSLS for training requires a lot of offline mining which incurs significant computational overhead, whereas training on \textsc{GSV-Cities} relies on online mining, which is much faster and requires no additional computation.

\subsection{Comparing different loss functions}
\label{sec:exp:metric_learning}
In this experiment, we carried out comparisons between five different metric learning loss functions for training place recognition networks: (1) Contrastive loss~\cite{hadsell2006dimensionality}; (2) Triplet loss~\cite{hermans2017defense}; (3) FastAP loss~\cite{cakir2019deep}; (4) Circle loss~\cite{sun2020circle}; and \mbox{(5) Multi-Similarity loss~\cite{wang2019multi}}. PyTorch implementation of these loss functions (and many others) can be found in~\cite{musgrave2020pytorch}. 
For training, we used a subset of $20$k places from \textsc{GSV-Cities} and trained the network for a maximum of $15$ epochs. We evaluated on two challenging benchmarks, Pitts30k-test~\cite{torii2013visual} and MSLS-val~\cite{warburg2020mapillary}.

As shown in Table \ref{tab:losses}, performance vary depending on the loss function used for the training. Multi-Similarity loss~\cite{wang2019multi} achieves the best results on both benchmarks, which might be explained by its sophisticated pair mining and weighting strategy. This shows the significance of the training loss function for place recognition networks. Notice how the contrastive loss improves to second best when coupled with a sophisticated mining strategy, which highlights the importance of online informative sample mining during the training. We believe \textsc{GSV-Cities} paves the way for new research into VPR-specific loss functions.

\setlength{\tabcolsep}{20pt}
\begin{table}[h]
\centering
\caption{Comparing different metric learning loss functions for training place recognition networks. MS-miner is the pair mining strategy used in Multi-Similarity loss, and OHM is a strategy that uses only the most difficult triplets in the batch. A subset of $20$k places from \textsc{GSV-Cities} was used for the training.}
\label{tab:losses}
\resizebox{0.75\textwidth}{!}{%
\begin{tabular}{l c@{\hspace{5pt}}c@{\hspace{5pt}}c c@{\hspace{5pt}}c@{\hspace{5pt}}c}
\toprule
\multirow{2}{*}{Loss function}                    & \multicolumn{3}{c}{Pitts30k-test}            & \multicolumn{3}{c}{MSLS-val}                 \\ 
                                                    \cmidrule(l{0.3cm} r{0.3cm}){2-4} \cmidrule(l{0.3cm} r{0.3cm}){5-7}
                                                  & \footnotesize{R@1}           & \footnotesize{R@5}           & \footnotesize{R@10}          & \footnotesize{R@1}           & \footnotesize{R@5}           & \footnotesize{R@10}          \\ \midrule
Contrastive \cite{hadsell2006dimensionality}            & 86.7          & 94.0          & 95.7          & 67.8          & 79.3          & 83.5          \\
Contrastive \cite{hadsell2006dimensionality} + MS-miner & 87.8          & 94.5          & 96.0          & 71.8          & 80.7          & 84.1          \\
Triplet  \cite{hermans2017defense} + OHM          & 85.2          & 93.0          & 95.1          & 60.4          & 72.6          & 76.1          \\
FastAP \cite{cakir2019deep}                       & 87.0          & 93.6          & 95.5          & 67.7          & 80.3          & 82.8          \\
Circle \cite{sun2020circle}                       & 86.9          & 94.3          & 95.8          & 72.2          & 82.8          & 85.5          \\
Multi-Similarity \cite{wang2019multi}             & \textbf{89.2} & \textbf{95.1} & \textbf{96.4} & \textbf{76.9} & \textbf{85.7} & \textbf{88.0} \\ \bottomrule
\end{tabular}
}
\end{table}

\subsection{Comparison to state-of-the-art}
\label{sec:exp:sota}
In this section, we compare the performance of our proposed aggregation method (Conv-AP) to existing techniques. We test four variants of Conv-AP$_{s \times s}$, by varying the size $s$ of the adaptive average pooling operation. For fair comparison, we train AVG~\cite{arandjelovic2016netvlad}, GeM~\cite{radenovic2018fine}, NetVLAD~\cite{arandjelovic2016netvlad}, SPE-VLAD~\cite{yu2019spatial}, Gated NetVLAD~\cite{zhang2021vector}, CosPlace~\cite{berton2022rethinking} and Conv-AP on \textsc{GSV-Cities} dataset, using the exact same configurations and hyperparameters.

\setlength{\tabcolsep}{10pt}
\begin{table*}[t]
\caption{Comparison with state-of-the-art approaches on $4$ place recognition benchmarks. Note that all models are trained on \textsc{GSV-Cities} and directly evaluated on Pitts250k, MSLS, SPED and Nordland datasets. Conv-AP$_{s \times s}$ is our method, with $s$ representing the height and width of the adaptive average pooling operation.}
\label{tab:sota-results}
\resizebox{\textwidth}{!}{%
\begin{tabular}{l c@{\hspace{5pt}}c@{\hspace{5pt}}c c@{\hspace{5pt}}c@{\hspace{5pt}}c c@{\hspace{5pt}}c@{\hspace{5pt}}c c@{\hspace{5pt}}c@{\hspace{5pt}}c }
\toprule
\multirow{2}{*}{Method}                & \multicolumn{3}{c}{Pitts250k-test}           & \multicolumn{3}{c}{MSLS-val}                 & \multicolumn{3}{c}{SPED}                     & \multicolumn{3}{c}{Nordland}                 \\ 
                                         \cmidrule(l{0.3cm} r{0.3cm}){2-4}              \cmidrule(l{0.3cm} r{0.3cm}){5-7}               \cmidrule(l{0.3cm} r{0.3cm}){8-10}               \cmidrule(l{0.3cm} r{0.3cm}){11-13}
                                       & \footnotesize{R@1}           & \footnotesize{R@5}           & \footnotesize{R@10}          & \footnotesize{R@1}           & \footnotesize{R@5}           & \footnotesize{R@10}          & \footnotesize{R@1}           & \footnotesize{R@5}           & \footnotesize{R@10}          & \footnotesize{R@1}           & \footnotesize{R@5}           & \footnotesize{R@10}          \\ \midrule
AVG~\cite{arandjelovic2016netvlad}     & 78.3          & 89.8          & 92.6          & 73.5          & 83.9          & 85.8          & 58.8          & 77.3          & 82.7          & 15.3          & 27.4          & 33.9          \\
GeM~\cite{radenovic2018fine}           & 82.9          & 92.1          & 94.3          & 76.5          & 85.7          & 88.2          & 64.6          & 79.4          & 83.5          & 20.8          & 33.3          & 40.0          \\
NetVLAD~\cite{arandjelovic2016netvlad} & 90.5          & 96.2          & 97.4          & 82.6          & 89.6          & 92.0          & 78.7          & 88.3          & 91.4          & 32.6          & 47.1          & 53.3          \\ 

SPE-VLAD~\cite{yu2019spatial} & 89.9          & 96.1          & 97.3          & 78.2          & 86.8          & 88.8          & 73.1          & 85.5          & 88.7          & 25.5          & 40.1          & 46.1          \\ 

Gated NetVLAD~\cite{zhang2021vector} & 89.7          & 95.9          & 97.1          & 82.0          & 88.9          & 91.4          & 75.6          & 87.1          & 90.8          & 34.4          & 50.4          & 57.7          \\ 

CosPlace~\cite{berton2022rethinking} & 91.5     & 96.9          & 97.9          & 83.0          & 89.9          & 91.8          & 75.3          & 85.9          & 88.6          & 34.4          & 49.9          & 56.5          \\

\midrule
Conv-AP$_{1 \times 1}$ (ours)          & 90.5          & 96.2          & 97.5          & 80.3          & 89.6          & 91.6          & 75.0          & 86.8          & 90.3          & 25.8          & 40.8          & 46.8          \\
Conv-AP$_{2 \times 2}$ (ours)          & \textbf{92.4} & 97.4          & 98.4          & \textbf{83.4} & \textbf{90.5} & \textbf{92.3} & 80.1          & 90.3          & 93.6          & \textbf{38.2} & \textbf{54.8} & \textbf{61.2} \\
Conv-AP$_{3 \times 3}$ (ours)          & 92.2 & \textbf{97.6} & \textbf{98.6}          &     83.2 & 89.2          & 91.1          & 80.9          & 90.3          & 93.4          & 34.8          & 50.1          & 56.2          \\
Conv-AP$_{4 \times 4}$ (ours)          & 92.2          & 97.4          & 98.3          & 80.1          & 88.6          & 90.3          & \textbf{81.2} & \textbf{90.3} & \textbf{93.9} & 34.3          & 48.6          & 55.8          \\ \bottomrule
\end{tabular}
}
\end{table*}

Experimental results are shown in Table~\ref{tab:sota-results}. As we can see, our method achieves substantially higher results than existing state-of-the-art on all benchmarks, reaching a new state-of-the-art on Pitts250k ($92.4\%$), MSLS ($83.4\%$), SPED ($81.2\%$) and Nordland ($38.2\%$).

Overall, Conv-AP$_{2 {\times} 2}$, which uses an adaptive average pooling window of size $2 {\times} 2$ (spatially splits the feature maps into four equally sub-regions), obtains the best results, beating NetVLAD and CosPlace on every benchmark. Furthermore, it is interesting to note that when $s=1$ (Conv-AP$_{1 \times 1}$) we see a relative performance drop, we believe this is due to the spatial dimension being collapsed to $1{\times}1$, potentially resulting in the loss of any spatial order present in the feature maps (see section~\ref{sec:exp:extended} for more experimental details).

Finally, we did not perform re-ranking of the top retrieved candidates as done in Patch-NetVLAD~\cite{hausler2021patch} and SuperGlue~\cite{sarlin2020superglue}. This is a technique known to boost recall@k by running a second matching pass that performs spatial verification of the local features. That being said, our method (Conv-AP) still outperforms Patch-NetVLAD and SuperGlue by a large margin. For instance, on MSLS-val, Conv-AP achieves recall@1 that is $3.9$ points higher than Patch-NetVLAD and $5$ points higher than SuperGlue. Moreover, on the Mapillary Challenge, Conv-AP outperforms Patch-NetVLAD by $10.4$ points ($68.0\%$ vs $57.6\%$ recall@5).

\subsection{Further analysis}\label{sec:exp:extended}
In this section, we investigate the robustness of our method with respect to its hyperparameters. Conv-AP produces representations of size $d {\times} s_1 {\times} s_2$ (that are flattened and L$_2$-normalized). Each parameter influences the size of the output and most likely performance. In Table~\ref{tab:output_sizes}, we conduct comprehensive experiments to show how the depth $d$ and the spatial size $s$ of Conv-AP affect performance. We note by Conv$_{d}$-AP$_{s {\times} s}$ each variant. 
We observe that reducing the depth $d$ does not necessarily result in lower performance. For instance, reducing the depth from $d=2048$ to $d=256$ produces $8$ times smaller representations with negligible effect on overall performance. This illustrates how Conv-AP can be configured to generate highly efficient representations without performance loss. For example, \mbox{Conv$_{512}$-AP$_{2 {\times} 2}$} outperforms NetVLAD while being $16\times$ more compact ($2048$-D vs $32768$-D). 
Moreover, \mbox{Conv$_{2048}$-AP$_{1{\times}1}$} and Conv$_{512}$-AP$_{2 \times 2}$ both produce $2048$-D representations. However, the latter outperforms the former on both benchmarks. This confirms our hypothesis in section~\ref{sec:exp:sota}, which states that collapsing the spatial dimension to~$1{\times}1$ may lead to a loss of spatial order in the resulting representations causing drop in performance.

\setlength{\tabcolsep}{15pt}
\begin{table}[tb]
\caption{Performance of Conv$_{d}$-AP$_{s \times s}$ where $d$ is the depth of the channel-wise pooling and $s$ is the size of the adaptive average pooling operation. ResNet50 is used  as backbone. Output dimension represents the size of the resulting flattened representations.}
\label{tab:output_sizes}
\centering
\resizebox{0.8\textwidth}{!}{%
\begin{tabular}{cc c c@{\hspace{5pt}}c@{\hspace{5pt}}c c@{\hspace{5pt}}c@{\hspace{5pt}}c }
\toprule
\multicolumn{2}{l}{Conv-AP hyperparams.}                        & \multirow{2}{*}{\begin{tabular}[c]{@{}c@{}}Output \\dimension\end{tabular}} & \multicolumn{3}{c}{Pitts250k-test} & \multicolumn{3}{c}{MSLS-val} \\ 
                                                                  \cmidrule(l{0.3cm} r{0.3cm}){1-2}                                             \cmidrule(l{0.3cm} r{0.3cm}){4-6}    \cmidrule(l{0.3cm} r{0.3cm}){7-9}
\multicolumn{1}{c}{$d$}    & $s_1 {\times} s_2$          &                                                                                & \footnotesize{R@1}           & \footnotesize{R@5}           & \footnotesize{R@10}      & \footnotesize{R@1}           & \footnotesize{R@5}           & \footnotesize{R@10}    \\ \midrule
\multicolumn{1}{r}{256}  & \multirow{4}{*}{$1 {\times} 1$} & 256                                                                         & 89.7       & 96.1       & 97.6      & 80.4     & 88.5     & 90.5    \\
\multicolumn{1}{r}{512}  &                               & 512                                                                            & 90.5       & 96.5       & 97.8      & 80.4     & 88.4     & 90.8    \\
\multicolumn{1}{r}{1024} &                               & 1024                                                                           & 90.4       & 96.0       & 97.4      & 81.8     & 89.7     & 90.9    \\
\multicolumn{1}{r}{2048} &                               & 2048                                                                           & 90.5       & 96.2       & 97.5      & 80.3     & 89.6     & 91.6    \\ \midrule
\multicolumn{1}{r}{256}  & \multirow{4}{*}{$2 {\times} 2$} & 1024                                                                           & 91.9       & 97.4       & 98.2      & 81.4     & 89.5     & 92.2    \\
\multicolumn{1}{r}{512}  &                               & 2048                                                                           & 92.0         & 97.4       & 98.2      & 82.7     & 90.4     & 92.8    \\
\multicolumn{1}{r}{1024} &                               & 4096                                                                           & 92.3       & 97.4       & 98.4      & 83.4     & 90.5     & 92.3    \\
\multicolumn{1}{r}{2048} &                               & 8192                                                                           & 92.4       & 97.4       & 98.4      & 82.2     & 89.1     & 90.9    \\ \bottomrule
\end{tabular}
}
\end{table}

\subsection{Backbone architectures}
\label{sec:exp:backbone}
\begin{figure}[h]%
\centering
\begin{subfigure}[-]{0.35\textwidth}
         \centering
         \includegraphics[width=\textwidth]{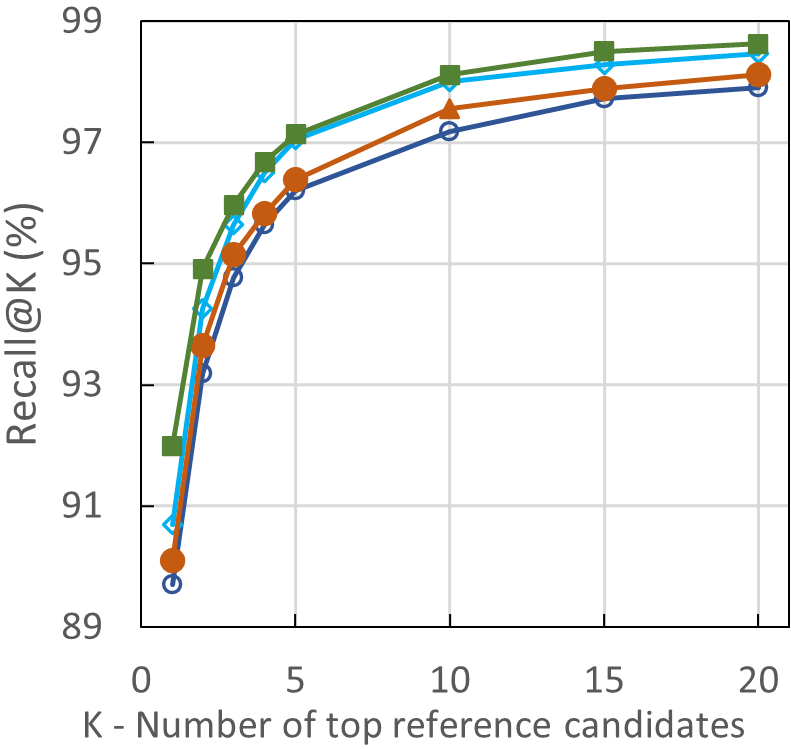}
         \caption{Pitts250k-test}
         \label{fig:backbone-pitts}
\end{subfigure}
\begin{subfigure}[-]{0.35\textwidth}
         \centering
         \includegraphics[width=\textwidth]{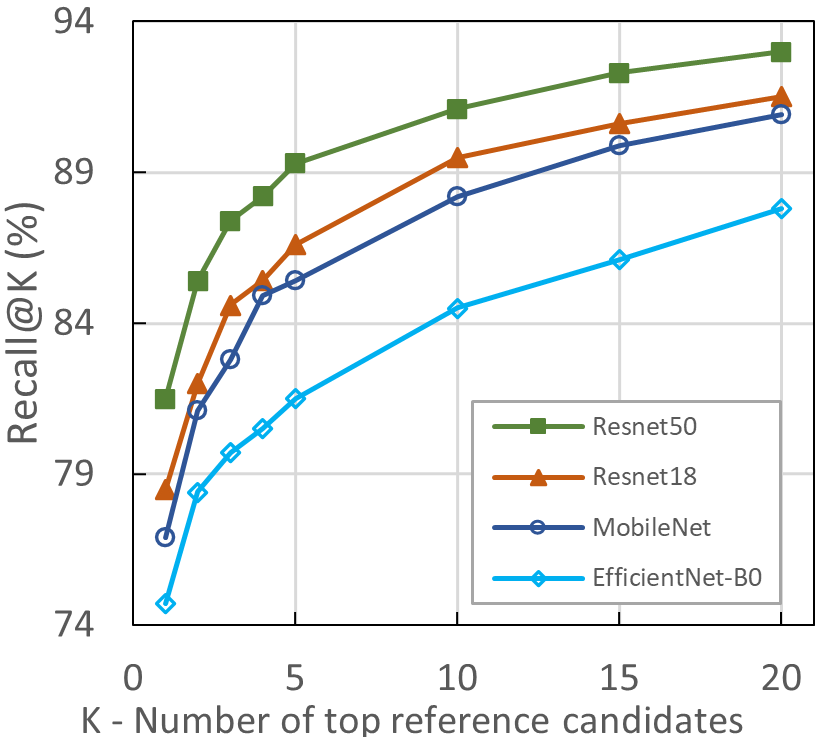}
         \caption{MSLS-val}
         \label{fig:backbone-MSLS}
\end{subfigure}
\caption{Performance of Conv-AP coupled with different backbone architectures. All models have been trained on \textsc{GSV-Cities}.}
\label{fig:backbone}
\end{figure}

We also compare between different backbone architectures, EfficientNet~\cite{tan2019efficientnet}, ResNet~\cite{he2016deep} and MobileNet~\cite{howard2019searching}. Fig.~\ref{fig:backbone} shows that compact backbones, such as ResNet18, MobileNet and EfficientNet-B0, can achieve top-notch performance on challenging benchmarks while requiring small memory footprint. ResNet50 obtains the best overall performance, especially on MSLS.

\subsection{Dimensionality reduction}\label{sec:exp:pca}
\begin{figure}[tb]
      \centering
      \includegraphics[scale=0.75]{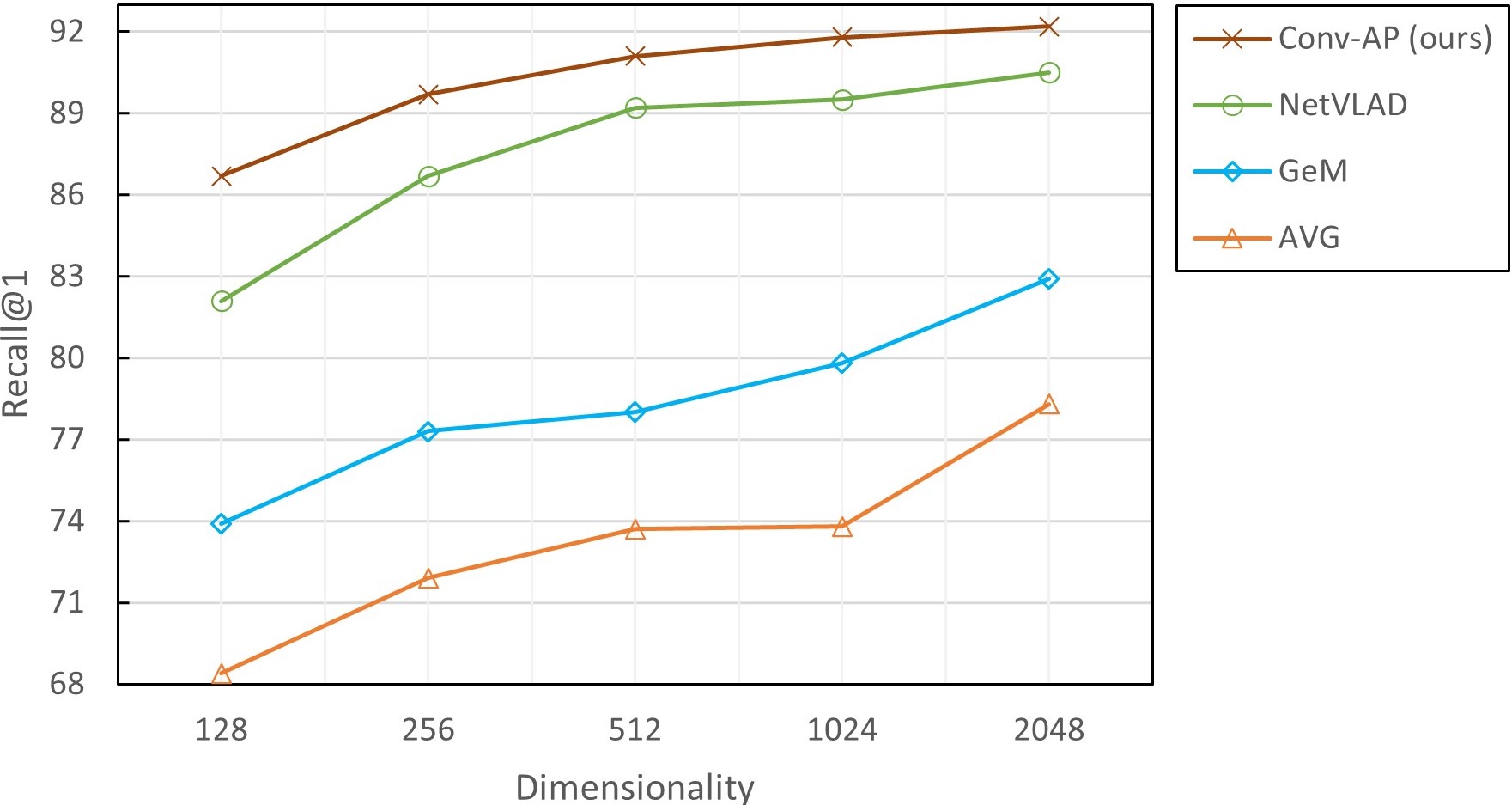}
  \caption{Recall@1 performance on Pitts250k-test after PCA dimensionality reduction. Note the log-scale of the x-axis. Conv-AP convincingly outperforms all other techniques. $512$-D Conv-AP performs better than $4\times$ larger $2048$-D NetVLAD.}
      \label{fig:PCA}
\end{figure}

Most state-of-the-art techniques use Principal Component Analysis (PCA) and Whitening~\cite{jegou2012negative} to reduce the dimension of the resulting representations in order to obtain compact image descriptors suitable for efficient storage. Although, our method (Conv-AP) can be configured to generate highly compact representations out of the box (e.g., by fixing $d=128$ and $s=2$, we obtain $512$-D outputs), we nevertheless perform PCA for fair comparison. All techniques are trained on \textsc{GSV-Cities}, and PCA is learned on a subset of $10$k images. Fig.~\ref{fig:PCA} shows recall@1 performance on Pitts250k-test. Our method outperforms all other techniques for any dimension size. For instance, $512$-D Conv-AP still outperforms $2048$-D NetVLAD while being $4\times$ more compact. Furthermore, these results, in direct agreement with those in Table~\ref{tab:output_sizes}, show that Conv-AP can generate highly efficient representations out of the box with minimal performance degradation.

\section{Conclusion}
In this paper, we introduced \textsc{GSV-Cities}, a large-scale dataset ($560$k images from $67$k locations) for appropriate training of visual place recognition methods. Importantly, the highly accurate ground truth of \textsc{GSV-Cities} eliminated the bottleneck of weak supervision that is currently limiting existing techniques, while improving their performance as well as drastically reducing training time.
Capitalizing on that, we showed that metric learning loss function can improve performance of VPR techniques when accurate labels are provided. We believe that this paves the way for further research into place recognition-specific architectures and loss functions.
Finally, we introduced Conv-AP, a fully convolutional aggregation method that significantly outperforms existing techniques. In this context, we established new state-of-the-art on the challenging Pitts250k-test, MSLS-val, SPED and Nordland benchmarks.

\vspace{10pt}
\noindent\textbf{Acknowledgement.} This work has been partially supported by Natural Sciences and Engineering Research Council of Canada (NSERC), The Fonds de Recherche du Québec Nature et technologies (FRQNT). We gratefully acknowledge the support of NVIDIA Corporation with the donation of a GPU used for our experiments.


\end{document}